\documentclass[letterpaper, 10 pt, journal, twoside]{ieeetran}

\IEEEoverridecommandlockouts                              

\usepackage{bm} % for using \bm command
\usepackage{multirow} % for tables
\usepackage{booktabs} % for table formatting
\usepackage{hyperref} % for adding links
\usepackage{subcaption} % for multifigures
\usepackage{caption} 
\usepackage{amsmath} % assumes amsmath package installed
\usepackage{amssymb}  % assumes amsmath package installed
\usepackage{xcolor} % to update colors of text

\title{Leading or Following? Dyadic Robot Imitative Interaction Using the Active Inference Framework}
\author{Nadine Wirkuttis$^{1}$ and Jun Tani$^{2,*}$%
\thanks{This work was sponsored by the Okinawa Institute of Science and Technology Graduate University, Okinawa, Japan 904-0302.}% 
\thanks{$^{1}$ The authors are with the Cognitive Neurorobotics Research Unit, Okinawa Institute of Science and Technology Graduate University, Okinawa, Japan 904-0302. {\tt\small \{nadine.wirkuttis,jun.tani\}@oist.jp}}%
\thanks{*Corresponding author.}%
}

\usepackage{breqn}
\usepackage{comment}
\begin{document}
\maketitle

% ABSTRACT
%
\begin{abstract}
This study investigated how social interaction among robotic agents changes dynamically depending on the individual belief of action intention. In a set of simulation studies, we examine dyadic imitative interactions of robots using a variational recurrent neural network model. The model is based on the free energy principle such that a pair of interacting robots find themselves in a loop, attempting to predict and infer each other's actions using active inference. We examined how regulating the complexity term to minimize free energy determines the dynamic characteristics of networks and interactions. When one robot trained with tighter regulation and another trained with looser regulation interact, the latter tends to lead the interaction by exerting stronger action intention, while the former tends to follow by adapting to its observations. The study confirms that the dyadic imitative interaction becomes successful by achieving a high synchronization rate when a leader and a follower are determined by developing action intentions with strong belief and weak belief, respectively.
\end{abstract}
%%%%%%%%%%%%%%%%%%%%%%%%%%%%%%%%%%%%%%%%%%%%%%%%%%%%%%%%%%%%%%%%%%%%%%%%%%%%%%%%
% INTRODUCTION
%
\section{INTRODUCTION}
\IEEEPARstart{S}{ocial} interaction is considered an essential cognitive behavior. In both empirical studies and synthetic modeling, researchers have investigated underlying cognitive, psychological, and neuronal mechanisms accounting for various aspects of social cognitive behaviors. 
This study investigates mechanisms underlying synchronized imitation as a representative social cognitive act, by formulating the problem using the free energy principle (FEP) \cite{friston2005theory,fristonMNS2011}. In simulation experiments of dyadic robot imitative interaction, we examine how a leader and follower can be determined in conflicting situations by investigating the underlying network dynamics.\par
Numerous robotic studies have investigated imitative interaction. In the 90s, imitation was identified as an indispensable human competency required in early development of cognitive behaviors \cite{Kuniyoshi1994, Hayes&Demiris1994, Dautenhahn1995, Gaussier1998}. Rizzolatti and colleagues \cite{Rizzolatti1992} showed that the mirror neuron system uses observations of an action to generate the same action. Arbib and Oztop \cite{Arbib2002,Oztop2006} indicated that mirror neurons may participate in imitative behaviors. Upon this development, several research groups proposed computational mirror neuron models for imitation using Hidden Markov Models \cite{Inamura2001} and neural network models \cite{Billard2001, Oztop2005, Ito&Tani2004, hwang2020dealing}.\par
An essential unsolved question in modeling of imitative interaction is how a leader, who initiates an action, and a follower, who imitates this action, can be determined when multiple choices of actions are possible among a set of well-habituated ones.\par
Recent theories on predictive coding (PC) and active inference (AIF) based on the free energy principle (FEP) \cite{fristonMNS2011,clark2015} show that "action intention" and its "belief" can be formulated as a predictive model. 
"Action intention" is considered a top-down prediction of action outcomes and "belief" as an estimated precision of this prediction or the strength of intention (as described in \cite{fristonMNS2011,clark2015}). 
Analogous to PC and AIF, Ito and Tani showed that imitative interaction can be performed using an RNN model by minimizing the prediction error instead of free energy in order to update deterministic latent variables \cite{Ito&Tani2004}. However, this deterministic model does not account for the belief of action intention because the precision of prediction cannot be estimated.
On a related topic, Ahmadi and Tani developed predictive-coding inspired variational RNN (PV-RNN) \cite{Reza2019}. Their model was used to investigate how the strength of top-down intention in predicting fluctuating temporal patterns was modulated, depending on learning conditions in the model. In the learning process, free energy represented by the weighted sum of the accuracy term and the complexity term is minimized. Ahmadi and Tani found that softer regulation of the complexity term during network training develops strong top-down intention. Predictions are more deterministic by self-organizing deterministic dynamics with the initial sensitivity characteristics in the network. Likewise, tighter regulation of the complexity term results in weaker intention and increased stochasticity. Compared to other neural network models based on the FEP \cite{Murata2015, Lanillos2018, Lang2018, Philippsen_Nagai2021}, PV-RNN has advantages when applied to problems in robotics. It can cope with temporal structure by using recurrence associated with stochastic latent variables and by hierarchical abstraction through a multiple timescale structure \cite{yamashita2008}.\par
Our research group further investigated human-robot imitative interaction using PV-RNN.
Chame and Tani \cite{chameICRA2020} showed that a humanoid robot with force feedback control tends to lead or follow the human counterpart in imitative interaction when its PV-RNN is set to softer or tighter regulation, respectively. 
However, the result is preliminary, merely showing a one-shot experimental result without any quantitative analysis. 
In a similar experimental setup, Ohata and Tani \cite{wataru2020} showed that this tendency can be also observed when regulation of the complexity term is modulated during the interaction phase, rather than during the prior learning phase.
The study investigated pseudo-imitative interaction between a simulated robot and a human counterpart. This study, however, lacks genuine interaction between the simulated robot and the human counterpart because the outputs of the counterpart were replaced with static output sequences prepared in advance.\par
The main contribution of the current study is to clarify the underlying mechanism of how a leader and a follower can be determined in dyadic synchronized imitative interaction using the framework of AIF.
This study is distinct from the author's aforementioned past studies because genuine interaction between two robots using the same model is examined and results are analysed both quantitatively and qualitatively.
An advantage of performing a robot-robot interaction experiment is that the internal dynamics can be analyzed in a comparative way between the two robots.\par
The interaction experiment considers two robotic agents that are trained to generate a set of movement primitive sequences. When movements are generated by following a probabilistic finite state machine, the transition probability differs, depending on each of the two robots. After each robot learns the given probabilistic transition structure for a sequence, the experimental design allows us to investigate how two robots generate movement primitives in the synchronised imitative interaction.
In particular, we examine conflicting situations in which each robot prefers to generate different movement patterns, depending on its learned experience.
Do they synchronize to generate the same movement pattern with one robot following the other or leading by adapting the intention? Or do they desynchronize by generating different movement patterns, ignoring their counterparts by following their own action intentions? The current study hypothesizes that these dyadic interaction outcomes depend on the relative strength of the intention between the robots as a result of regulating FEP complexity.
%
%
%%%%%%%%%%%%%%%%%%%%%%%%%%%%%%%%%%%%%%%%%%%%%%%%%%%%%%%%%%%%%%%%%%%%%%%%%%%%%%%%%
% MODEL
%
\section{Model}\label{sec:model}
\subsection{Predictive Coding and Active Inference}
The current study applies the concepts of PC and AIF based on FEP \cite{friston2005theory}. PC considers perception as the interplay between a \textit{prior} expectation of a sensation and a \textit{posterior} inference about a sensory outcome. Expectation of the sensation can be modeled by a generative model that maps the prior of the latent state to the expectation of sensation. 
The posterior inference of the observed sensation can be achieved by jointly minimizing the error computed between the expected sensation and its outcome, i.e. the {\it accuracy}, plus the Kullback-Leibler (KL) Divergence between the posterior and the prior distributions, i.e. the {\it complexity}. Posterior and prior are both represented by Gaussian probability distributions using means and variances. This is to minimize free energy or to maximize the lower bound of the logarithm of marginal likelihood: 
\begin{equation}
\begin{aligned}\footnotesize
    \log p_\theta(\bar{\boldsymbol{X}})&\geq\underbrace{\int q_\phi(\boldsymbol{z}|\bar{\boldsymbol{X}})\log\frac{p_\theta(\bar{\boldsymbol{X}},\boldsymbol{z})}{q_\phi(\boldsymbol{z}|\bar{\boldsymbol{X}})}d\boldsymbol{z}}_{\rm Evidence\ lower\ bound}\\
    &=\underbrace{\mathbb{E}_{q_\phi(\boldsymbol{z|\bar{\boldsymbol{X}}})}[\log p_\theta(\bar{\boldsymbol{X}}|\boldsymbol{z})]}_{\rm Accuracy}-\underbrace{D_{\rm KL}[q_\phi(\boldsymbol{z}|\bar{\boldsymbol{X}})\Vert p_\theta(\boldsymbol{z})]}_{\rm Complexity}\label{eq:elbo}
\end{aligned}
\end{equation}
$\boldsymbol{z}$, $\bar{\boldsymbol{X}}$, $p_\theta$, and $q_\phi$ denote the latent state, the observation, the prior distribution, and the approximate posterior, respectively. $\theta$ and $\phi$ are the parameters of the generative and inference model. 
In maximizing the lower bound, the interplay between {\it accuracy} and {\it complexity} characterizes the model performance in learning, prediction, and inference.\par
Consistent with the AIF, actions are generated so that the error between the expected action outcome and the actual outcome is minimized. In robotic applications, this is equivalent to determining how expected proprioception in terms of robot joint angles can be achieved by generating adequate motor torque. A simple solution is to use a PID controller, in which adequate motor torque to minimize errors between expected joint angles and actual angles can be obtained by means of error feedback schemes. 
Finally, perception by predictive coding and action generation by active inference are deployed simultaneously, thereby closing the loop of action and perception.
%
% PVRNN
%
\subsection{Overview of PV-RNN}
The PV-RNN model is designed to predict future sensation by means of prior generation, while reflecting the past by means of posterior inference based on learning (see Fig. \ref{fig:architecture}). 
One essential element of the model is the introduction of a parameter $\mathbf{w}$, the so-called meta-prior, which regulates the complexity term in free energy. Different $\mathbf{w}$ settings results in alternation of the estimated precision in predicting the sensation, as described later as \textit{prior generation} (see section \ref{subsec:experiments_standalone}). The model is also characterised by employing an architecture of multiple timescale RNN (MTRNN) \cite{yamashita2008}. The whole network comprises multiple layers of RNNs wherein the dynamics of each layer are governed by different time constant parameters $\tau$. 
This scheme supports development of hierarchical information processing by adequately setting the timescale of each layer \cite{yamashita2008, hwang2020dealing}. 
This approach is considered as analogous to \cite{Pio-lopez-hierarchy-AIF2016,Schillaci2020}.\par
The following briefly describes the two essential parts, a \textit{generative model} which is used for prior generation to make future predictions, and an \textit{inference model}, which is used for posterior inference about the past. For further details, refer to \cite{Reza2019,wataru2020}.
%
% generative model
%
\subsubsection{Generative Model}
\begin{figure}[thpb]
  \centering
  \includegraphics[width=0.45\textwidth]{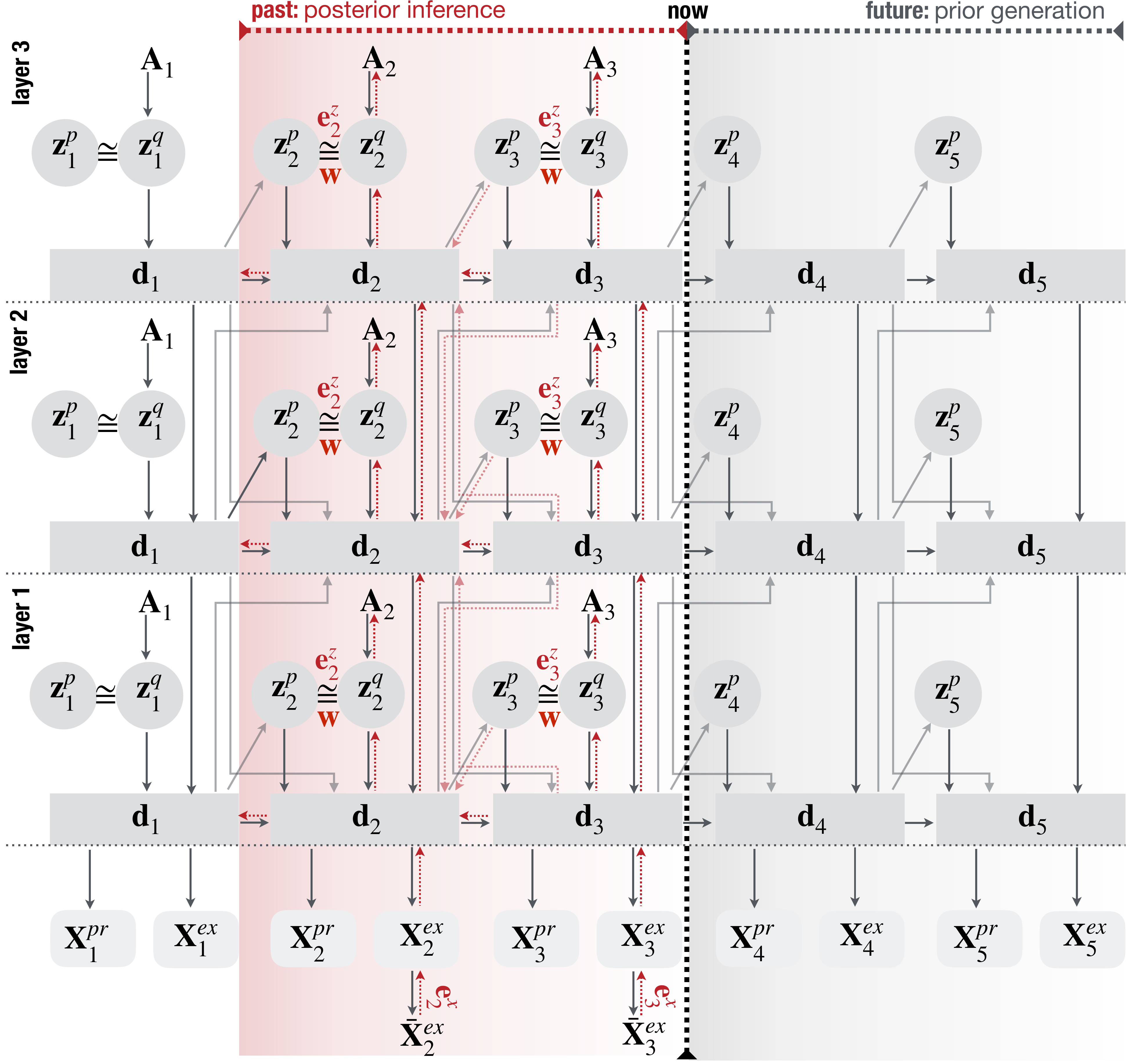}
  \caption{\textbf{Graphical representation of a hierarchical 3-layer PV-RNN architecture.} Layers are indicated on the left. Time increases from left to right and is indicated as a subscript. The representation shows the network in $t=3$ with a two-time-step posterior inference window $[2,3]$ and prior generation for $t=[4,5]$. In the posterior inference window, prediction error $\mathbf{e}^x$ and the $\mathbf{w}$ weighted KL Divergence $\mathbf{e}^z$ are minimized.}
  \label{fig:architecture}
\end{figure}

The stochastic generative model is used for prior generation, as illustrated in the future prediction part (after time step 4) in Fig. \ref{fig:architecture}.
PV-RNN is comprised of deterministic variables $\mathbf{d}$ and random variables $\mathbf{z}$. An approximate posterior distribution $q_\phi$ is inferred based on the prior distribution $p_\theta$ by means of error minimization on the generated prediction $\mathbf{X}$. The generative model can be factorized as:

\begin{equation}
\begin{aligned}
\label{eq:genp} \footnotesize
    p_\theta(\bar{\bm{X}}_{1:T}, {\mathbf{d}}_{1:T}, \bm{z}_{1:T} | \bm{d}_0) = \\ \prod_{t=1}^T p_{\theta_{\bar{\bm{X}}}}(\bar{\bm{X}}_t | {\mathbf{d}}_t)  p_{\theta_{{d}}}({\mathbf{d}}_t | {\mathbf{d}}_{t-1}, \bm{z}_t) p_{\theta_z}(\bm{z}_t | {\mathbf{d}}_{t-1})
\end{aligned} 
\end{equation} 
Although $\mathbf{d}$ is a deterministic variable, it can be considered to have a Dirac delta distribution centered on $\tilde{\bm{d}}$ as $\bm{\sigma}(\bm{d}-{\tilde{\bm{d}}})$. $\bar{\bm{X}}$ is conditioned directly on $\bm{z}$ through $\tilde{\bm{d}}$. At the initial time step, $\tilde{\bm{d}}$ is set to $0$. Otherwise, $\tilde{\bm{d}}$ is recursively computed, for which the internal state before activation is denoted by $\bm{h}$. 
This internal state $\bm{h}$ is a vector, calculated as the sum of the internal states of the current level $l$ and its connecting layers of the previous time step $t-1$ plus the latent $\bm{z}$ in the same layer of the current time step $t$:
\begin{equation} \label{eq:cell} \footnotesize
\begin{aligned}
    \bm{\tilde{d}}^l_t &= \text{tanh}(\bm{h}^l_t)\\
    \bm{h}^l_t &= \left(1 - \frac{1}{\tau^l}\right)\bm{h}^l_{t-1} +
    \\&\frac{1}{\tau^l} \left( \bm{W}^{ll}_{dd}\bm{\tilde{d}}^l_{t-1} + \bm{W}^{ll}_{zd}\bm{z}^l_t + \bm{W}^{ll+1}_{dd}\bm{\tilde{d}}^{l+1}_{t-1} + \bm{W}^{ll-1}_{dd}\bm{\tilde{d}}^{l-1}_{t-1} \right)
\end{aligned}
\end{equation}
 $\tau^l$ denotes the layer-specific time constant. With larger value for $\tau^l$, slower timescale dynamics develop, whereas with a smaller value set, faster timescale dynamics dominate.
 $\bm{W}$ represents connectivity weight matrices between layers and their deterministic and stochastic units. 
 The output with size $N_x$ is computed as mapping from $\bm{\tilde{d}}^{1}$ as:
\begin{equation} \label{eq:output} \footnotesize
\begin{aligned}
\mathbf{X}_t = \bm{W}^{ll}_{Xd}\bm{\tilde{d}}_t^1 + \bm{b}_X
\end{aligned}
\end{equation}
 The prior distribution $p_\theta(\bm{z}_t)$ is a Gaussian distribution represented with mean $\bm{\mu}_t^p$ and standard deviation $\bm{\sigma}_t^p$. The prior depends on $\bm{\tilde{d}}_{t-1}$ by following the idea of a sequence prior \cite{chung2015recurrent}, except at $t=1$ where it follows a unit Gaussian distribution.
\begin{equation} \label{eq:p} \footnotesize
\begin{aligned}
    p_\theta(\bm{z}_1) &= \mathcal{N}(0, I)\\
    p_\theta(\bm{z}_t | \tilde{\bm{d}}_{t-1}) &= \mathcal{N}(\bm{\mu}^p_t, (\bm{\sigma}^p_t)^2) \text{ where $t>1$}\\
    \bm{\mu}^p_t &= \text{tanh}(\bm{W}^{ll}_{d\mu^p}\bm{\tilde{d}}_{t-1})\\ 
    \bm{\sigma}^p_t &= \exp(\bm{W}^{ll}_{d\sigma^p}\bm{\tilde{d}}_{t-1})
\end{aligned}
\end{equation}

Based on the work on variational autoencoders, we use the \textit{reparameterization trick} to formulate the latent prior of $\bm{z}_t$ as mean $\bm{\mu}^p_t$ and standard deviation $\bm{\sigma}^p_t$. The \textit{reparameterization trick} was introduced by Kingma and Welling \cite{kingma2014autoencoding} to make random variables differentiable for backpropagating errors through the network for learning. The same consideration is taken for the posterior of $\bm{z}_t$ in the inference model as well (cf. below Eq. \ref{eq:q}). 
%
% inference model
%
\subsubsection{Inference Model}
Posterior inference is performed during learning and afterward, during action and perception. Fig. \ref{fig:architecture} illustrates information flow in the posterior inference in a time window from time step 2 to time step 3. 
The inference model for the posterior is described as:
\begin{equation} 
\begin{aligned}
\label{eq:Q}\footnotesize
Q_\phi(\bm{z}_t | \bm{\tilde{d}}_{t-1}, \bm{e}_{t:T}) = \mathcal{N}(\bm{z}_t; \bm{\mu}^q_t, \bm{\sigma}^q_t)
\end{aligned}
\end{equation}
where $\bm{e}_{t}$ denotes the error between the target $\bar{\mathbf{X}}_t$ and the predicted output $\mathbf{X}_t$. Like the prior $p_\theta$, the posterior $q_\phi$ is also a Gaussian distribution with mean $\bm{\mu}_t^q$ and standard deviation $\bm{\sigma}_t^q$. For $\bm{z}_{1:T}$ it is defined as:
\begin{equation} \label{eq:q} \footnotesize
\begin{aligned}
    q_\phi(\bm{z}_t | \bm{e}_{t:T}) &= \mathcal{N}(\bm{\mu}^q_t, \bm{\sigma}^q_t)\\
    \bm{\mu}^q_t &= \text{tanh}(\bm{A}^\mu_t)\\
    \bm{\sigma}^q_t &= \exp(\bm{A}^\sigma_t)
\end{aligned}
\end{equation}
Since computing the true posterior is intractable, an approximate posterior $q_\phi$ is inferred by maximizing the lower bound, analogous to Eq. (\ref{eq:elbo}). Here, the adaptation variable $\mathbf{A}_{1:T}$ forces the parameters $\phi$ of the inference model to represent meaningful information.
The lower bound of PV-RNN can be derived as:
\begin{equation} 
\begin{aligned}\footnotesize
    L(\theta, \phi) 
    & = \sum_{t=1}^T (\frac{1}{N_{X}} E_{q_\phi(\bm{z}_t | \bm{\tilde{d}}_{t-1}, \bm{e}_{t:T})} \big[\log  p_{\theta_{\bar{X}}}(\bar{\bm{X}}_t | \bm{\tilde{d}}_{t}) \big] - \\
    & \sum_{l}^L \frac{\mathbf{w}^l}{N_{z}^l} D_{KL}\big[ q_\phi(\bm{z}_t | \bm{\tilde{d}}_{t-1}, \bm{e}_{t:T}) || p_{\theta_z} (\bm{z}_t | \bm{\tilde{d}}_{t-1}) \big])
\label{eq:femin}
\end{aligned}
\end{equation}
where the first term is the \textit{accuracy} and the second term is the \textit{complexity} (for details referred to \cite{Reza2019}). $N_x$ and $N_z^l$ are the number of sensory dimensions and the number of the latent random variables at the $l^{th}$ layer, respectively. $\mathbf{w}^l$ serves as a weighting parameter for the complexity term in layer $l$ and is referred to as the meta-prior \cite{Reza2019}. 
The meta-prior represents the strength for regulating the closeness between the posterior and the prior distributions.
In $t=1$, $\mathbf{w}^l_1$ is set with 1.0. $\mathbf{w}^l_{2:T}$ is set to a specific value when the sequence prior \cite{chung2015recurrent} is used after time step 1. In the posterior inference, all learning-related network parameters of $\theta$, $\phi$, and the adaptive variable $\mathbf{A}$ are updated to maximize the lower bound by back-propagating the error from time step $T$ back to $t_1$ \cite{rumelhart1985learning}. 
\subsubsection{PV-RNN in Dyadic Robot Interaction}
\begin{figure}[thpb]
  \centering
  \includegraphics[trim=0 20 0 -10mm, width=0.40\textwidth]{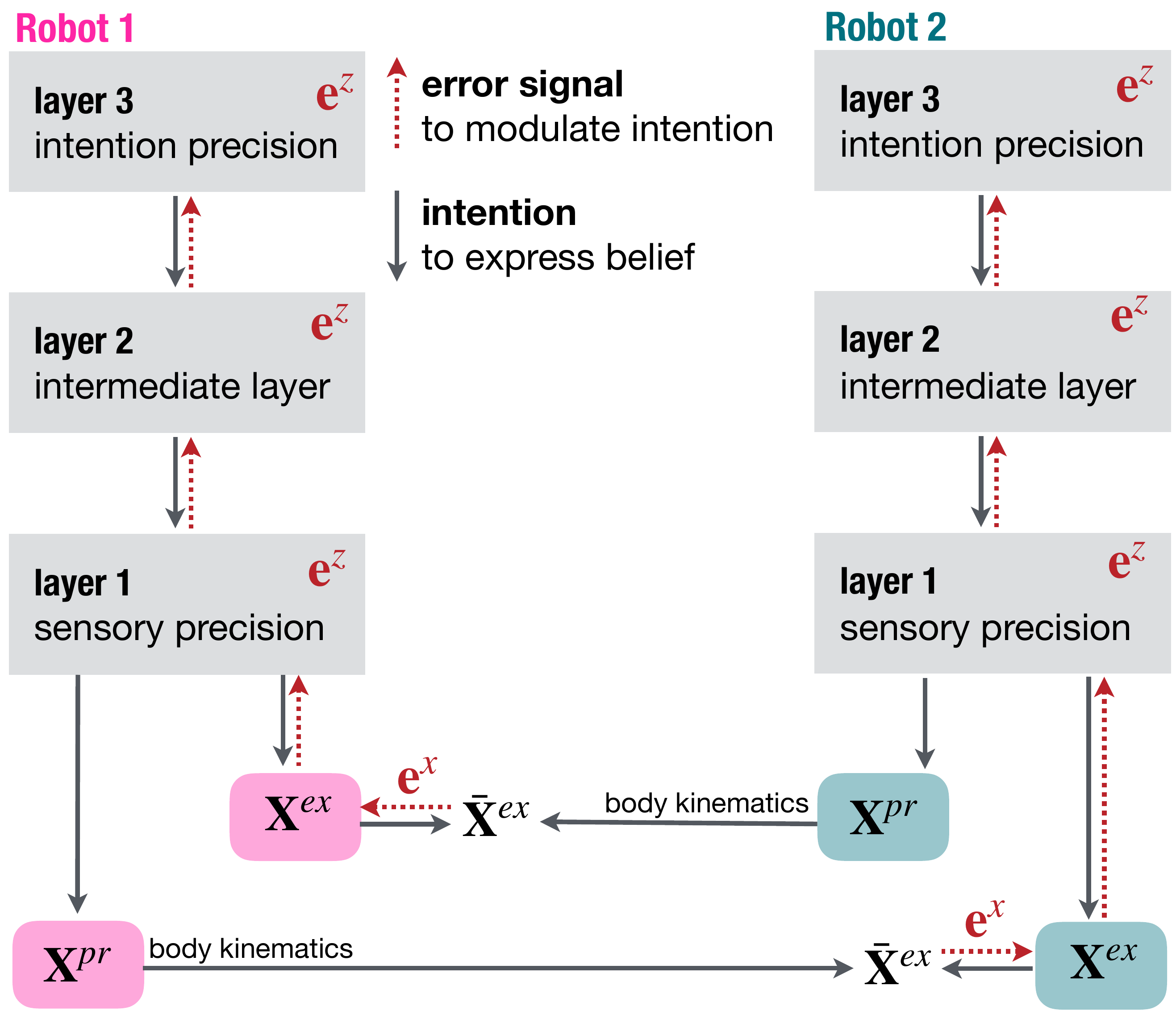}
  \caption {\textbf{Schematic of dyadic robot interaction} where robots are equipped with the PV-RNN model.}
  \label{fig:interaction}
\vspace{-4mm}
\end{figure}

Two robots equipped with the PV-RNN model interact during synchronized imitation. In the interaction, the robots predict proprioception $\mathbf{X}^{pr}_{t+1}$ and exteroception $\mathbf{X}^{ex}_{t+1}$ for the next time step. The predicted $\mathbf{X}^{pr}_{t+1}$ regulates joint angle movements of a robot by considering a PID controller. This movement $\mathbf{X}^{pr}_{t+1}$ can then be sensed by the other robot in terms of exteroception $\bar{\mathbf{X}}^{ex}_{t+1}$. This is provided through the kinematic transformation of joint angles $\mathbf{X}^{pr}_{t+1}$ (cf. Fig. \ref{fig:interaction}). 
While in the training phase, the error signal is taken from the proprioceptive $\bar{\mathbf{X}}^{pr}$ as well as the exteroceptive $\bar{\mathbf{X}}^{ex}$ target sequences, in the interaction phase the error signal for each robot is taken only from $\bar{\mathbf{X}}^{ex}$\footnote{Considering only $\bar{\mathbf{X}}^{ex}$ for the error term in interaction settings assumes that the PID controller generates only negligible position errors for the joints.}. During interactions, prediction errors ${\mathbf{e}}^{x}$ are generated and propagated bottom-up throughout all layers, as well as time steps in the posterior inference window, in terms of the latent error ${\mathbf{e}}^{z}$. This updates posterior distributions in the network and minimizes the variational free energy. In this phase, only $\mathbf{A}_{1:T}$ is updated without updating network parameters $\theta$ and $\phi$.%
%
%%%%%%%%%%%%%%%%%%%%%%%%%%%%%%%%%%%%%%%%%%%%%%%%%%%%%%%%%%%%%%%%%%%%%%%%%%%%%%%%%
%
\section{Robot Experiments}\label{sec:experiments}
To investigate how the interaction of two robots changes with tighter and looser regulation of complexity, each robot was trained and tested individually, as described in \ref{subsubsec:training} and in \ref{subsec:experiments_standalone}, respectively. Finally, two robots were examined during a dyadic interaction (\ref{subsec:experiments_dyad}).
\subsection{Task Design}\label{subsubsec:dataset}
\begin{figure}
\center
\begin{subfigure}{.45\textwidth}
    \centering
    \includegraphics[trim=0 0 0 -5mm, width=\textwidth]{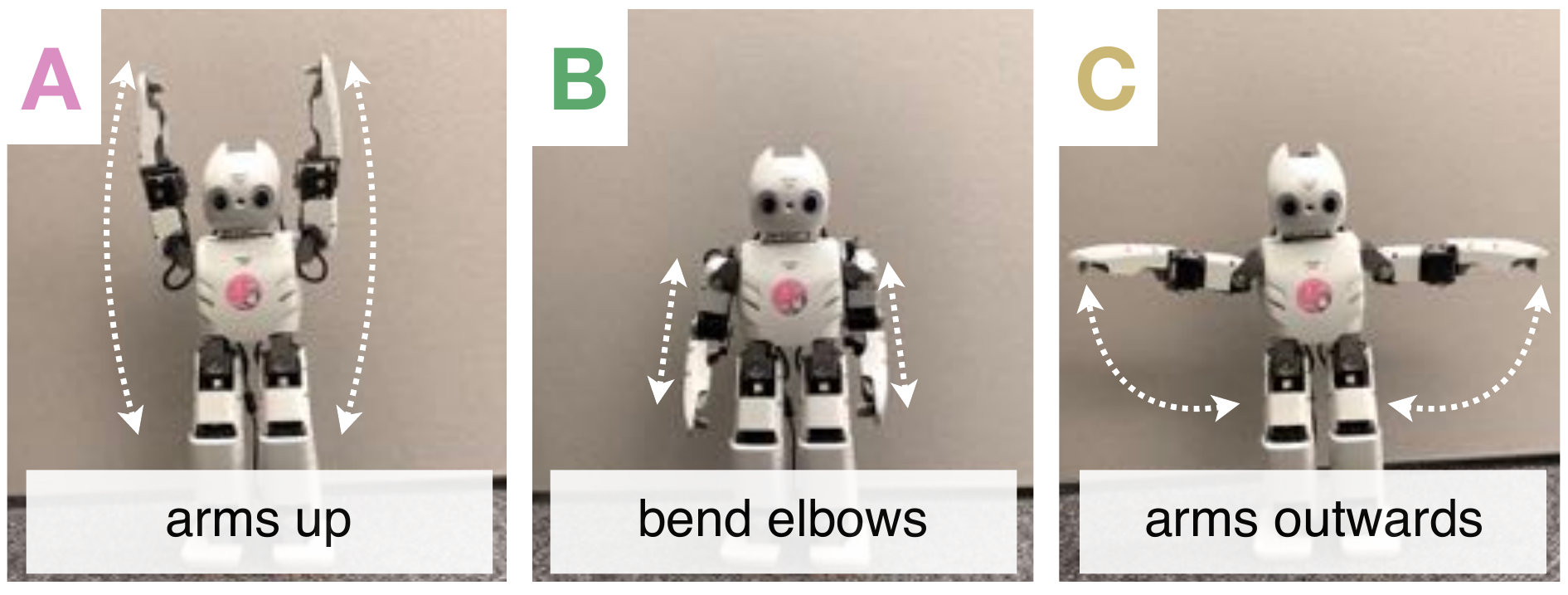}
    \caption{robot movements}
    \label{subfig:dataset_primitives}
\end{subfigure}
%\hfill
\begin{subfigure}{.45\textwidth}
    \centering
    \includegraphics[width=\textwidth]{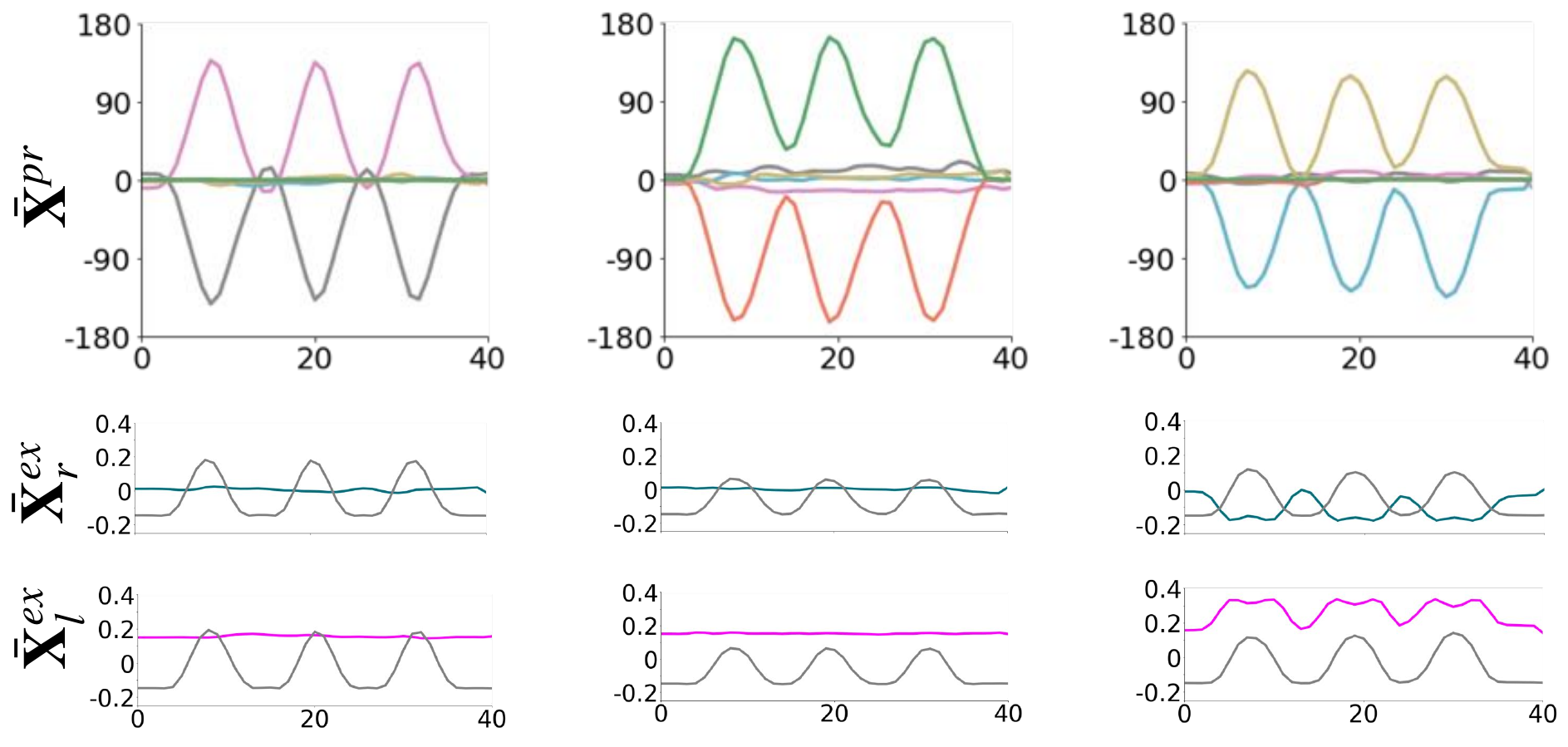}
    \caption{movement trajectories}
    \label{subfig:dataset_trajectories}
\end{subfigure}
%\hfill
\begin{subfigure}{.45\textwidth}
    \centering
    \includegraphics[width=\textwidth]{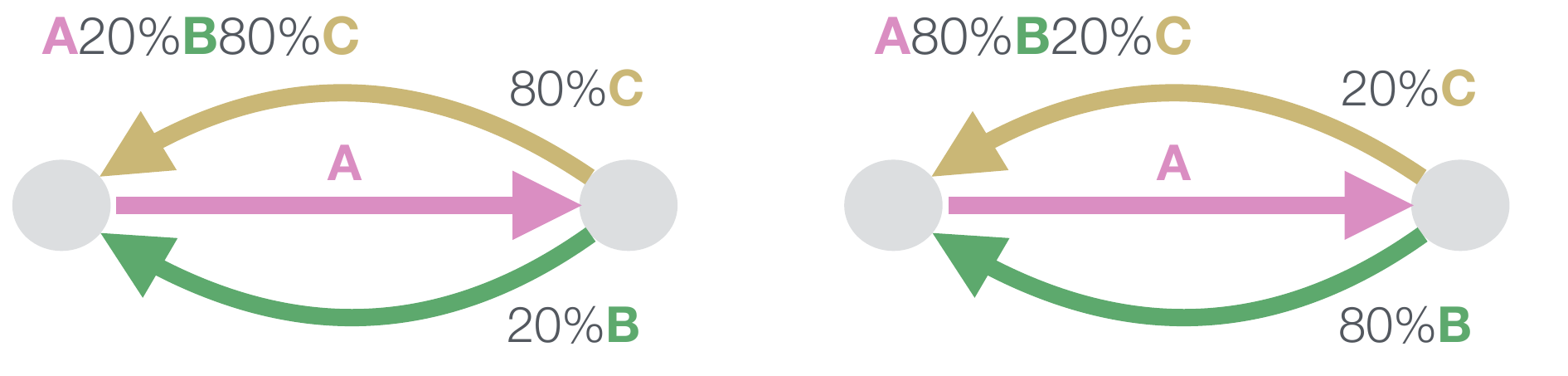}
    \caption{probabilistic transition}
    \label{subfig:dataset_switching}
\end{subfigure} 
\caption{\textbf{Task design.} Robot movement primitives {\sffamily\textbf{A}}, {\sffamily\textbf{B}}, and {\sffamily\textbf{C}} of the training dataset (a). 
Proprioceptive trajectories $\bar{\mathbf{X}}^{pr}$ plus exteroceptive trajectories $\bar{\mathbf{X}}^{ex}_{r}$ and $\bar{\mathbf{X}}^{ex}_{l}$. Colors represent six dimensions of joint angles for $\bar{\mathbf{X}}^{pr}$ and xy-coordinate positions of the right and left end effector $\bar{\mathbf{X}}^{ex}_r$ and $\bar{\mathbf{X}}^{ex}_{l}$ (b). Two P-FSMs representing different movement primitive transition patterns of
{\sffamily\textbf{A}}20\%{\sffamily\textbf{B}}80\%{\sffamily\textbf{C}} and {\sffamily\textbf{A}}80\%{\sffamily\textbf{B}}20\%{\sffamily\textbf{C}} (c).}
\label{fig:dataset}
\vspace{-4mm}
\end{figure}

Robotic agents are trained with three movement primitives {\sffamily\textbf{A}}, {\sffamily\textbf{B}}, and {\sffamily\textbf{C}} (Fig. \ref{fig:dataset} (a)). Each primitive is 40 time steps in length. A human experimenter generated the primitive data via a master control of a humanoid OP2\footnote{Humanoid OP2 and its master controller are developed by Robotis: \href{www.robotis.us/robotis-op2-us/}{www.robotis.us/robotis-op2-us/}.}. The experimenter controlled six joints in the upper body of one humanoid $\bar{{\mathbf{X}}}^{pr}$. The exteroceptive trajectory $\bar{{\mathbf{X}}}^{ex}$ is generated by mirroring its own movement $\bar{{\mathbf{X}}}^{pr}$ and transformed into $\bar{{\mathbf{X}}}^{ex}$ xy-coordinate positions of the left hand and right hand tips of the robot (Fig. \ref{subfig:dataset_trajectories}).
$\bar{{\mathbf{X}}}^{pr}$ and $\bar{{\mathbf{X}}}^{ex}$ are six and four dimensions, respectively. Individual movement primitives are sampled and combined to form a continuous pattern of 400 time steps that follows a probabilistic sequence (analogous to \cite{wataru2020}). Two probabilistic patterns were generated, {\sffamily\textbf{A}}20\%{\sffamily\textbf{B}}80\%{\sffamily\textbf{C}} and {\sffamily\textbf{A}}80\%{\sffamily\textbf{B}}20\%{\sffamily\textbf{C}} as shown in the form of a probabilistic finite state machine (P-FSM) (Fig. \ref{fig:dataset} (c)). The difference between these two probabilistic patterns is that {\sffamily\textbf{C}} is biased and comes more often (80\%) than {\sffamily\textbf{B}} (20\%) after {\sffamily\textbf{A}} in the former, and vice versa for the latter.\par
A point of interest is the interaction phase after the learning phase. It is expected that both robots can generate {\sffamily\textbf{A}} synchronously, since it is a deterministic state. This could be different from generating {\sffamily\textbf{B}} or {\sffamily\textbf{C}} as two robots learned different preferences in terms of transition probabilities. One robot may lead so as to generate {\sffamily\textbf{B}} or {\sffamily\textbf{C}} while the other may just follow it. However, both robots may generate their own biased movements and, thus, desynchronize their behavior.
The current study hypothesizes that whether {\sffamily\textbf{B}} or {\sffamily\textbf{C}} is generated synchronized or desynchronized between the two robots depends on the complexity regulation of each robot.
\subsection{Robot Training}\label{subsubsec:training}
The PV-RNN was trained with 20 data samples on a set of different parameters (TABLE \ref{tab:parameters}). All network specific parameters were fixed during training. To explore the influence of the meta-prior, w, only this parameter changed for different networks and was repeated with different random seeds to ensure reproducibility. 
%
%TABLE: network parameters
%
\begin{table}[htbp]\centering
\caption{\textbf{Network parameters} for training PV-RNNs.}
\label{tab:parameters}
\begin{tabular}{lccccc}
\toprule\footnotesize
     & $\mathbf{d}$ & $\mathbf{z}$ & $\tau$ & $\mathbf{w}_{1}$ & $\mathbf{w}_{2:T}$ \\
\cmidrule(lr){2-6}
\textbf{layer 1} & 40 & 4 & $2$ & $1$ & $\mathbf{w}^1=[0.0, 0.001, ... 4.999, 5.0]$ \\
\textbf{layer 2} & 20 & 2 & $4$ & $1$ & $\mathbf{w}^2={w}^1 \times 10$ \\
\textbf{layer 3} & 10 & 1 & $8$ & $1$ & $\mathbf{w}^3={w}^1 \times 100$ \\
\bottomrule
\end{tabular}
\vspace{-4mm}
\end{table}
Networks were trained for 80,000 epochs, using Adam Optimizing and back-propagation through time (BPTT) \cite{rumelhart1985learning} with learning rate 0.001. After training, network performance was first analysed in stand-alone robot experiments (subsection \ref{subsec:experiments_standalone}). Thereafter, dyadic robot interaction was studied using networks trained with $\mathbf{w}$ set for the two representatives of tight and loose regulation of FEP complexity (subsection \ref{subsec:experiments_dyad}).
%
% STAND ALONE EXPERIMENTS
%
\subsection{Preparatory Analysis of Training Results}\label{subsec:experiments_standalone}
%
% table training performance
%
\begin{table*}[tb]
\centering\caption{\textbf{Training performance} of representative meta-prior $\mathbf{w}$. The mean $\pm$ standard deviation represent three random seeds and 20 repetitions of prior generation for each $\mathbf{w}$.}
\label{tab:performance}
\begin{tabular}{lcccccc}
\toprule\footnotesize
%
% A20B80C
$\mathbf{w}$ & \multicolumn{4}{c}{\textbf{training sequence} {\sffamily\textbf{A}}\textbf{20}\%{\sffamily\textbf{B}}\textbf{80}\%{\sffamily\textbf{C}}} & 
{\sffamily\textbf{B}}{\sffamily\textbf{C}}\textbf{-ratio} &
\textbf{divergence step $\mathbf{t}$} \\ 
\cmidrule(lr){2-5} 
    & {\sffamily\textbf{A}} & {\sffamily\textbf{B}}      & 
    {\sffamily\textbf{C}}  &  not classified  & & \\    
\midrule
$0.005$   &  
$34\pm1$  & 
$11\pm3$ & 
$40\pm2$ & 
$15\pm1$ & 
$22\pm6${\sffamily\textbf{B}} $78\pm5${\sffamily\textbf{C}} & 
$43$\\ 
$0.01$   &  
$35\pm2$  & 
$13\pm0.2$ & 
$30\pm2$ & 
$22\pm4$ & 
$30\pm5${\sffamily\textbf{B}} $70\pm5${\sffamily\textbf{C}} &
$50$ \\
$1.0$   &  
$36\pm1$  & 
$11\pm2$ & 
$40\pm2$ & 
$13\pm0.2$ & 
$22\pm4${\sffamily\textbf{B}} $78\pm4${\sffamily\textbf{C}} &
$91$ \\
$2$   &  
$41\pm0.7$  & 
$10\pm0.5$ & 
$39\pm0.7$ & 
$10\pm0.3$ & 
$21\pm8${\sffamily\textbf{B}} $79\pm8${\sffamily\textbf{C}} &
$120$ \\
$3.4$   &  
$45\pm1$  & $11\pm1$ & $38\pm2$ & $6\pm0.5$ & 
$24\pm2${\sffamily\textbf{B}} $76\pm2${\sffamily\textbf{C}} &
$139$\\
\bottomrule
\end{tabular}
\vspace{-4mm}
\end{table*}
To investigate how the model learns the probabilistic structure of the training data, we conducted a first analysis in the form of \textit{prior regeneration}.
For prior regeneration we choose one training sample and use two time steps of the adaption variable $\mathbf{A}_{1:2}^{\bar{\mathbf{X}}}$ to initialize the prior distribution $p(\mathbf{z}_{1:2})$ in the PV-RNN. Thereafter the future prediction ${\mathbf{X}}_{3:400}$ for the remaining training sample length can be calculated (cf. \textit{prior generation} in Fig. \ref{fig:architecture}).
Using this scheme, we generated 20 sequences for each meta-prior $\mathbf{w}$. This was repeated for each network that was trained for that parameter for all random seeds. For brevity, training analysis is reported only for the network that was trained on the probabilistic sequence {\sffamily\textbf{A}}20\%{\sffamily\textbf{B}}80\%{\sffamily\textbf{C}}. Training of {\sffamily\textbf{A}}80\%{\sffamily\textbf{B}}20\%{\sffamily\textbf{C}} showed comparable results. An Echo State Network for multivariate time series classification \cite{bianchi2020reservoir} with reservoir size $N=45$, $25\%$ connectivity and leakage $60\%$ was used for classification of movement primitives. Movement patterns were identified as \textit{not classified} below a $55\%$ threshold.

\subsubsection{Analysis of Probabilistic Transition}
A robot that is trained with {\sffamily\textbf{A}}$20\%${\sffamily\textbf{B}}$80\%${\sffamily\textbf{C}} will first generate an {\sffamily\textbf{A}} movement, and then transition to {\sffamily\textbf{B}} with 20 percent probability and to {\sffamily\textbf{C}} with 80 percentage probability.
We found that smaller $\mathbf{w}$ settings are less stable in reproducing the probabilistic structure of the training data. 
The {\sffamily\textbf{B}}{\sffamily\textbf{C}}-ratio was either greater or less than $20\%$ for {\sffamily\textbf{B}} or greater or less than $80\%$ for {\sffamily\textbf{C}}. Networks trained with larger meta-priors become more reliable in regenerating the probabilistic training sequence ({\sffamily\textbf{B}}{\sffamily\textbf{C}}-ratio in Table \ref{tab:performance}). In addition to the capacity of learning the probability distribution of the training data, we found that smaller meta-priors show noisier pattern generation. Non-classified movements were as high as $22\%\pm4$ with $\mathbf{w}=0.01$ and decreased to $6\%\pm0.6$ with $\mathbf{w}=3.4$. 
\subsubsection{Divergence Analysis}
Repeatability in generating sequences in prior generation was examined by conducting a divergence analysis. Sequences are considered diverged when a comparison per time step of $\mathbf{X}^{pr}$ exceeds a threshold\footnote{As the threshold for the divergence analysis, we use the mean squared error of joint angle data $[-180, 180]$ of $\mathbf{X}^{pr}$. The threshold is set to $55$.}. Out of 20 regeneration sequences, we randomly select one as a reference and calculate the average divergence step of the other sequences to this the reference. Out of 400 time steps of prior generation, sequences diverged from the reference around time step $43$ for networks trained with smaller $\mathbf{w}$. With increasing $\mathbf{w}$, repeatability of the trajectories increased. Here the divergence step was around $139$ (cf. divergence step $t$ in Table \ref{tab:performance}).
\subsubsection{Summary of Preparatory Analysis}
Loose regulation of the complexity term results in noisier, less repeatable prior generation performance. Also the learned probability for transition to either {\sffamily\textbf{B}} or {\sffamily\textbf{C}} is not accurate. This observation changes with increasing meta-prior. The larger $\mathbf{w}$, the more accurate the learned transition probability becomes. Also, prior generation becomes more repeatable by developing more deterministic dynamics with the initial sensitivity characteristics (i.e., the sequence is generated solely depending on the latent state in the initial time step). For subsequent dyadic robot interaction experiments, we empirically select the meta-prior setting $\mathbf{w}=0.005$ and $\mathbf{w}=3.4$ as two representatives of tight and loose regulation of the FEP complexity.
%
% DYADIC EXPERIMENTS
%
\subsection{Dyadic Robot Interaction Experiments}\label{subsec:experiments_dyad} 
In the following experiments, robots are either trained with $\mathbf{w}=0.005$ or $\mathbf{w}=3.4$. For readability, we will consider {$\mathbf{R}^1_w$} and {$\mathbf{R}^2_w$} with subscripts of the respective meta-priors $\mathbf{w}$.
In the dyadic interaction, we present the network of each robot with observations of movements of the counterpart robot $\bar{\mathbf{X}}^{ex}$ as the target and perform posterior inference in a regression window with size $win_{size}=70$. Inference is performed from the current time step $t$ back to $t-win_{size}$, or $t_1$ in case $t-win_{size}\leq1$.
After 200 epochs of iteration to maximize the lower bound, the time window is shifted one time step forward. Note, all experiments were conducted in simulation due to the difficulty of real-time posterior inference computation.\par
We investigated how two robots interact in three different dyadic conditions (TABLE \ref{tab:interaction_performance}). We then analysed whether the robots trained with  {\sffamily\textbf{A}}80\%{\sffamily\textbf{B}}20\%{\sffamily\textbf{C}} maintained the learned preference between {\sffamily\textbf{B}} and {\sffamily\textbf{C}} or adapted to their counterparts that were trained with {\sffamily\textbf{A}}20\%{\sffamily\textbf{B}}80\%{\sffamily\textbf{C}}. We also calculated the so-called {\sffamily\textbf{B}}{\sffamily\textbf{C}}-synchronization rate during the interaction. If at any time step $t$, one of the robots generated {\sffamily\textbf{B}} or {\sffamily\textbf{C}} and the other robot generated the same movement primitive, the interaction was considered synchronized. 
Note that time steps in which movement patterns were identified as \textit{not classified} by the Echo State Network (cf. subsection \ref{subsec:experiments_standalone}) were excluded from the computation.\par
\begin{figure}[p]
\center
\begin{subfigure}{.47\textwidth}
    \centering
    \includegraphics[trim=0 0 0 -5mm,width=\textwidth]{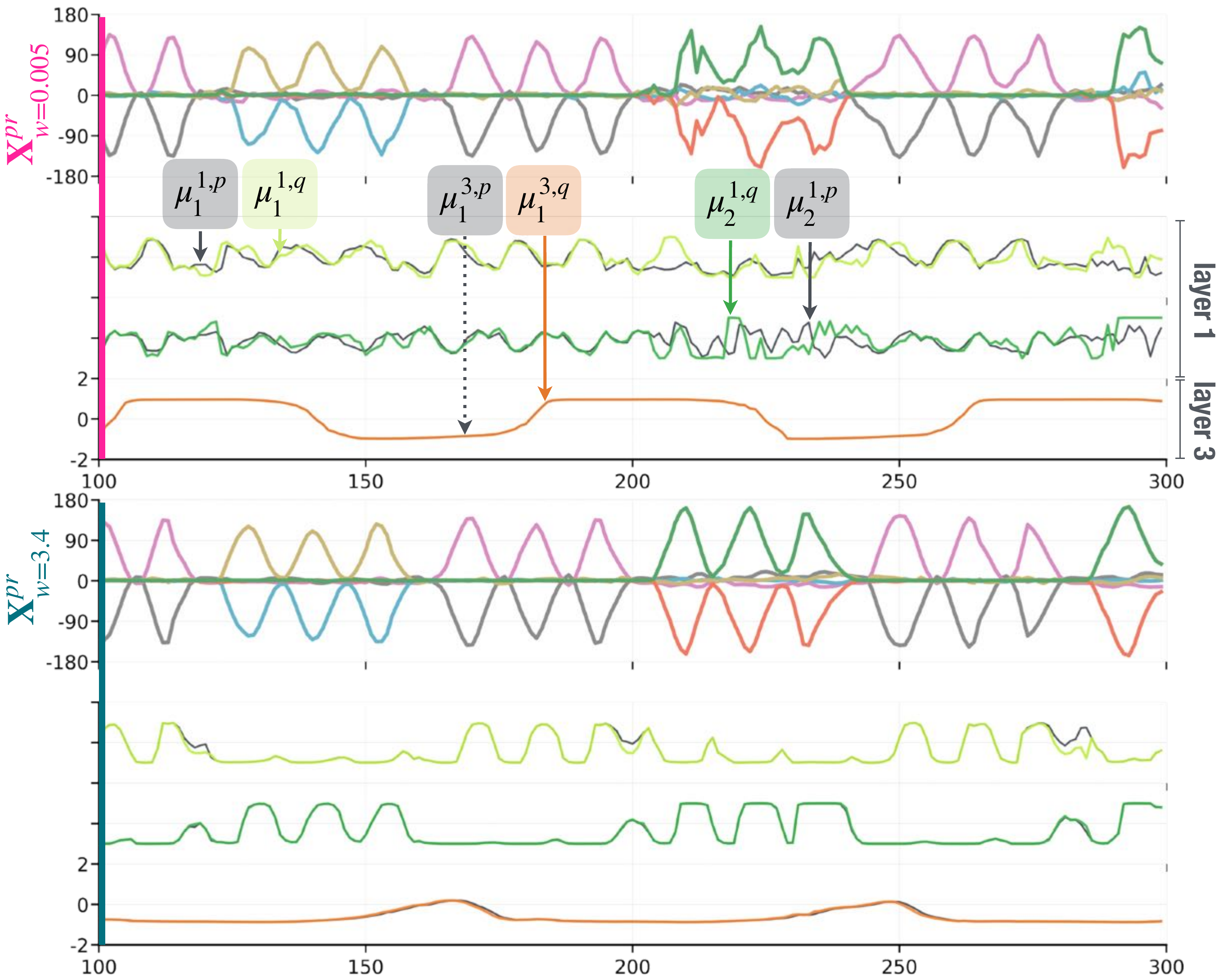}
    \caption{Experiment 1: {$\mathbf{R}^1_{0.005}$} vs {$\mathbf{R}^2_{3.4}$} }
    \label{subfig:interaction_weak-strong}
\end{subfigure}
\begin{subfigure}{.46\textwidth}
    \centering
    \includegraphics[width=\textwidth]{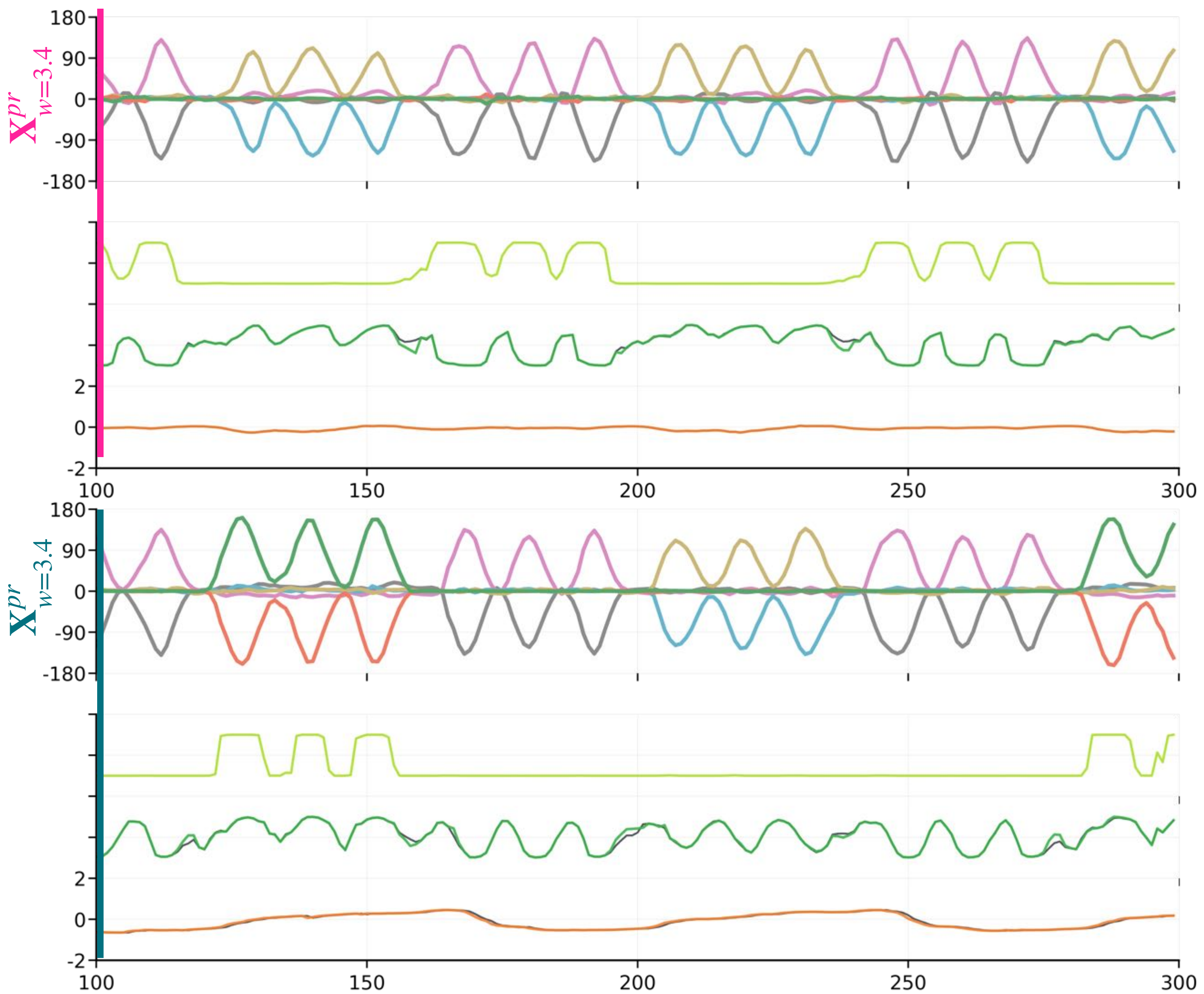}
    \caption{Experiment 2: {$\mathbf{R}^1_{3.4}$} vs {$\mathbf{R}^2_{3.4}$} }
    \label{subfig:interaction_strong-strong}
\end{subfigure}
\begin{subfigure}{.46\textwidth}
    \centering
    \includegraphics[width=\textwidth]{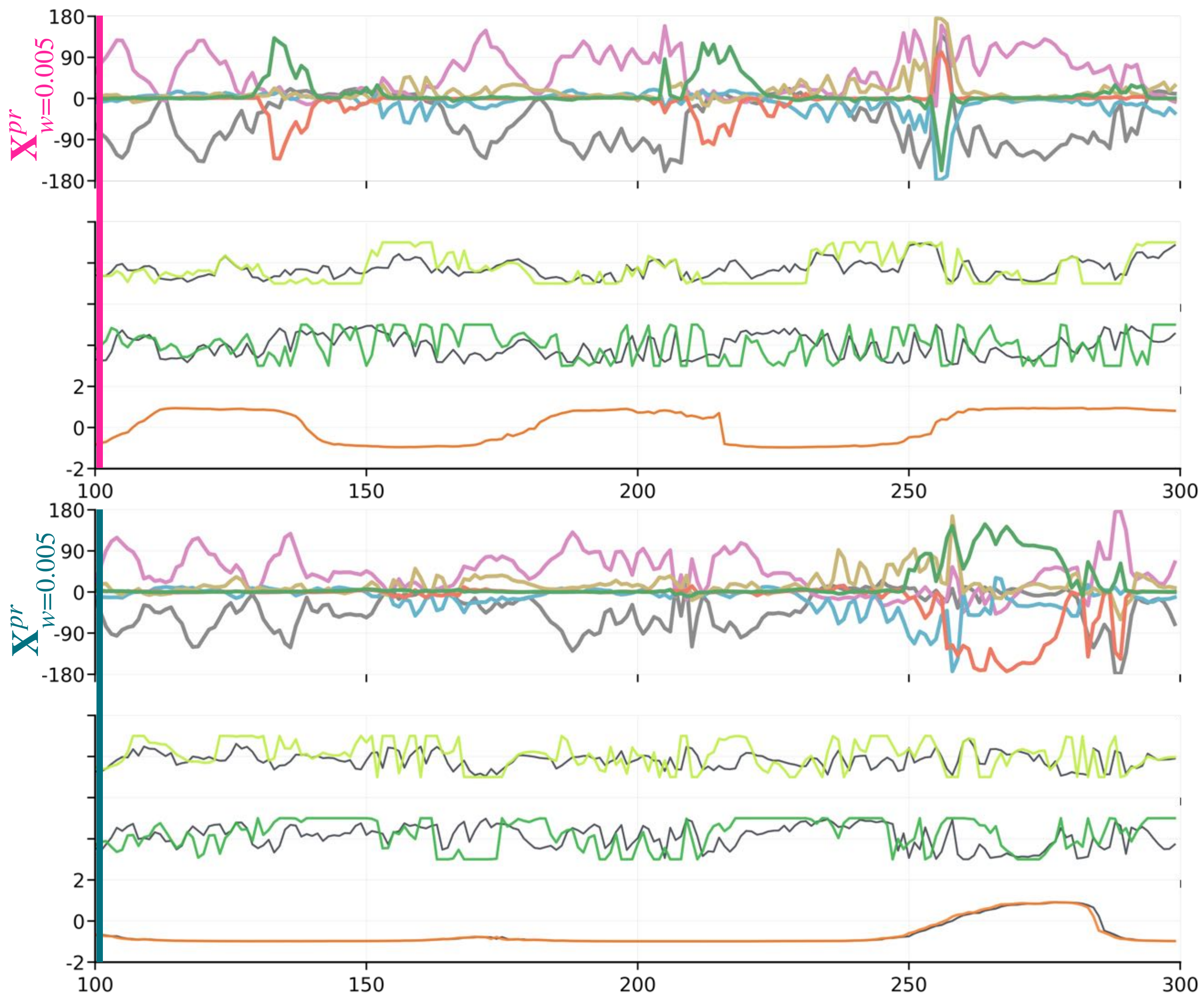}
    \caption{Experiment 3: {$\mathbf{R}^1_{0.005}$} vs {$\mathbf{R}^2_{0.005}$} }
    \label{subfig:interaction_weak-weak}
\end{subfigure}
\caption{\textbf{Movement trajectories and network dynamics of dyadic robot interaction} for Experiment 1 (a), Experiment 2 (b) and Experiment 3 (c). Time steps $t=[100,300]$ of movements and selected neurons in layer 1 and 3 are shown.}
\label{fig:experiments_dyad}
\vspace{-4mm}
\end{figure}

 
%
% REVISED TABLE %%%%%%%%%%%%%%%%%%%%%%%%%%%%%%%
\begin{table}[thpb]
\centering\caption{\textbf{Interaction performance} of three experimental settings.}
\label{tab:interaction_performance}
\begin{tabular}{llccccc}%\footnotesize
\toprule
\multicolumn{2}{c}{\textbf{Experiment}} & 
\multicolumn{4}{c}{{\sffamily\textbf{B}}{\sffamily\textbf{C}}\textbf{-ratio}} & 
{\sffamily\textbf{B}}{\sffamily\textbf{C}}-\textbf{sync} \\\cmidrule(lr){1-2}\cmidrule(lr){3-6}
ID & robots & \multicolumn{2}{c}{stand-alone} & \multicolumn{2}{c}{interaction} & \\\cmidrule(lr){3-4}\cmidrule(lr){5-6}
  &  & {\sffamily\textbf{B}} & {\sffamily\textbf{C}} & 
  {\sffamily\textbf{B}} & {\sffamily\textbf{C}} & \\
\midrule
%
% Experiment 1
\multirow{2}{*}{\textbf{1}} & {$\mathbf{R}^1_{0.005}$}&
$22\pm6$ & $78\pm5$  & 
$70\pm11$ & $30\pm10$  & 
\multirow{2}{*}{$56\pm23$} \\
& {$\mathbf{R}^2_{3.4}$}& 
$75\pm2$ & $25\pm3$ &
$73\pm10$ & $27\pm10$  & \\
\midrule
% Experiment 2
\multirow{2}{*}{\textbf{2}} & {$\mathbf{R}^1_{3.4}$}  &
$24\pm2$ & $76\pm2$  & 
$17\pm12$  & $83\pm18$   & 
\multirow{2}{*}{$31\pm24$} \\
& {$\mathbf{R}^2_{3.4}$} & 
$75\pm2$ & $25\pm3$ & 
$61\pm12$& $39\pm12$& \\
\midrule
% Experiment 3
\multirow{2}{*}{\textbf{3}} & {$\mathbf{R}^1_{0.005}$} &
$22\pm6$ & $78\pm5$ & 
$52\pm13$ & $48\pm9$ & 
\multirow{2}{*}{$42\pm20$}\\
& {$\mathbf{R}^2_{0.005}$} & 
$44\pm11$ & $56\pm11$ &
$20\pm7$ & $80\pm8$ & \\
\bottomrule
\end{tabular}
\vspace{-4mm}
\end{table}
TABLE \ref{tab:interaction_performance} shows the summary of the analysis for all three experiments. To better understand effects of loose and tight regulation of FEP complexity, exemplar plots of robot movement patterns, as well as corresponding network dynamics, are shown (cf. Fig. \ref{fig:experiments_dyad} and Fig. \ref{fig:interaction_weak-strong_prediction}). We provide supplementary movies of the experiments showing humanoid robot interaction and network dynamics here: \href{https://doi.org/10.6084/m9.figshare.14099537}{https://doi.org/10.6084/m9.figshare.14099537}.
\subsubsection{Experiment 1: {$\mathbf{R}^1_{0.005}$} vs. {$\mathbf{R}^2_{3.4}$}}\label{subsubsec:ex1}
In Experiment 1, {$\mathbf{R}^1_{0.005}$} adapts to the probabilistic transition of {$\mathbf{R}^2_{3.4}$} by increasing the probability of performing {\sffamily\textbf{B}} from $22\%$ in the stand-alone condition to $70\%$ in the dyad (Table \ref{tab:interaction_performance} Experiment 1). Both robots are performing more {\sffamily\textbf{B}} than {\sffamily\textbf{C}} with a {\sffamily\textbf{B}}{\sffamily\textbf{C}}-synchronization of $56\pm23\%$ which is significantly higher than the chance rate of $32 \%$\footnote{We assume that {\sffamily\textbf{B}} and {\sffamily\textbf{C}} are independent probabilistic events. Then we can consider the probabilities for a robot $R$ to perform either a {\sffamily\textbf{B}} movement as $\mathbf{P}^R(B)$ or a {\sffamily\textbf{C}} movement as $\mathbf{P}^R(C)$. The actual {\sffamily\textbf{B}}{\sffamily\textbf{C}}-synchronization chance level can then be calculated as: $\mathbf{P}^1(B) \times \mathbf{P}^2(B) + \mathbf{P}^1(C) \times \mathbf{P}^2(C) = 0.8 \times 0.2 + 0.8 \times 0.2 = 0.16 + 0.16 = 0.32$.}.\par
\begin{figure}[tbhp]
\center
\includegraphics[trim=0 10 0 -5mm,width=.48\textwidth]{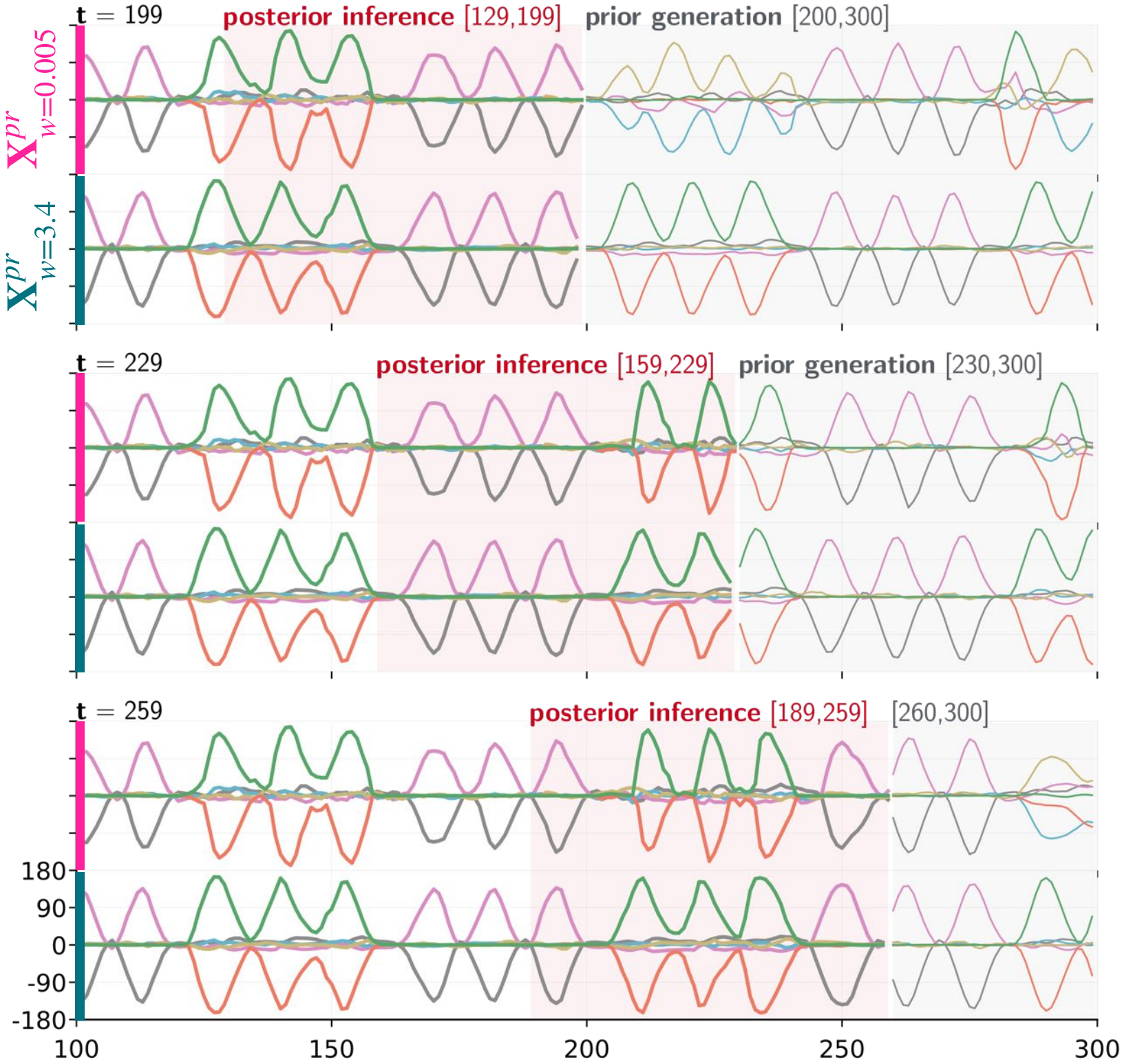}
\caption{\textbf{Posterior inference and prior generation in Experiment 1}. 
Interaction of {$\mathbf{R}^1_{0.005}$} (upper) and {$\mathbf{R}^2_{3.4}$} (lower) in terms of $\mathbf{X}^{pr}$. The first, the second, and the third row show $\mathbf{X}^{pr}$ after the posterior inference in the inference window with size $win_{size}=70$, as well as its future prior generation with current time steps of 199, 229, and 259, respectively.
}
\label{fig:interaction_weak-strong_prediction}
\end{figure}

Fig. \ref{fig:interaction_weak-strong_prediction} shows an example of how prediction of the future and posterior inference of the past proceed as time passes from time step $199$, $229$, to $259$ for both robots. We observe that the intended future behavior (the prior generation) of {$\mathbf{R}^1_{0.005}$} is not consistent with the actually performed actions after posterior inference. On the other hand, in the case of {$\mathbf{R}^2_{3.4}$}, the performed action complies with its prediction.
This behavior can be explained by looking at exemplar priors $\mu
_i^{lp}$ and posteriors $\mu_i^{lq}$ for layer $l$ and neuron $i$ between two robots. In layer $1$, selected posterior network dynamics $\mu_1^{1q}$ and $\mu_2^{1q}$ are deviating from prior dynamics $\mu_1^{1p}$ and $\mu_2^{1p}$ for robot {$\mathbf{R}^1_{0.005}$}. Whereas the dynamics of {$\mathbf{R}^2_{3.4}$} are mostly overlapping (cf. Fig. \ref{subfig:interaction_weak-strong} and \href{https://doi.org/10.6084/m9.figshare.14099537}{supplementary movie}). 
More specifically, the average KL Divergence $\mathbf{e}^{z}$ of {$\mathbf{R}^1_{0.005}$} is larger for all layers ($({\mathbf{e}}^{z,1},{\mathbf{e}}^{z,2},{\mathbf{e}}^{z,3})=(109.1, 1.4, 0.06)$) than for {$\mathbf{R}^2_{3.4}$} ($({\mathbf{e}}^{z,1},{\mathbf{e}}^{z,2},{\mathbf{e}}^{z,3})=(0.4, 0.0003, 0.00001)$). This means that {$\mathbf{R}^2_{3.4}$} tends to behave as intended because the posterior is attracted by the prior. On the other hand, {$\mathbf{R}^1_{0.005}$} tends to adapt to {$\mathbf{R}^2_{3.4}$} since the posterior is rather attracted by the observation than by the weaker prior belief.\par 
Note that $\mu_1^{3q}$ and $\mu_1^{3p}$ in layer 3 change only slowly with time. This indicates that these latent variables represent how movement primitives transit from deterministic states to non-deterministic states using their slower timescale properties characterized by $\tau^3$.
%
% EX2
%
\subsubsection{Experiment 2: {$\mathbf{R}^1_{3.4}$} vs. {$\mathbf{R}^2_{3.4}$}}\label{subsubsec:ex2}
When two robots with loose complexity regulation interact, both robots maintain their learned preferences in terms of probability in generating either {\sffamily{\textbf{B}}} or {\sffamily\textbf{C}}.
{$\mathbf{R}^1_{3.4}$}, which learns a $76\%$ transition to {\sffamily\textbf{C}} in a stand-alone situation, shows its preference to {\sffamily\textbf{C}} in the dyad with probability of $83\%$.
{$\mathbf{R}^2_{3.4}$}, which in a stand-alone condition would maintain its preference to {\sffamily\textbf{B}} with a probability of $75\%$, shows $61\%$ percentage transition to {\sffamily\textbf{B}} in the interaction.
{\sffamily\textbf{B}}{\sffamily\textbf{C}}-synchronization rate turns out to be low as $31\pm24\%$, which is almost equal to the chance rate. Examining the network dynamics of the prior and posterior distributions shows that the robots executed movements based upon their prior action intention without adapting their posteriors to observations of the other robot’s movement (cf. Fig. \ref{subfig:interaction_strong-strong} and \href{https://doi.org/10.6084/m9.figshare.14099537}{supplementary movie}). 
\subsubsection{Experiment 3: {$\mathbf{R}^1_{0.005}$} vs. {$\mathbf{R}^2_{0.005}$}}\label{subsubsec:ex3}
When two robots with tight regulation of complexity interact, both try to adapt their own action to the one demonstrated by the other. Indeed, Fig. \ref{subfig:interaction_weak-weak} shows that the prior and posterior do not comply, but deviate. Whether trained with the probabilistic transition of {\sffamily\textbf{A}}20\%{\sffamily\textbf{B}}80\%{\sffamily\textbf{C}} or {\sffamily\textbf{A}}80\%{\sffamily\textbf{B}}20\%{\sffamily\textbf{C}}, both robots significantly reduce the tendency to perform their own intended behavior  {\sffamily\textbf{C}} or {\sffamily\textbf{B}}, respectively. This is evidenced by changes of the {\sffamily\textbf{B}}{\sffamily\textbf{C}}-ratio from stand-alone compared to the dyadic setting (TABLE \ref{tab:interaction_performance} Experiment 3). {\sffamily\textbf{B}}{\sffamily\textbf{C}}-synchronization rate is $42\pm20\%$ which is higher than the chance rate but not significantly. The interaction becomes noisier, compared to results of Experiments 1 and 2 (cf. Fig. \ref{subfig:interaction_weak-weak} and \href{https://doi.org/10.6084/m9.figshare.14099537}{supplementary movie}), which indicate that tight regulation makes robots more sensitive to temporal fluctuations in observations of their counterparts.
%%%%%%%%%%%%%%%%%%%%%%%%%%%%%%%%%%%%%%%%%%%%%%%%%%%%%%%%%%%%%%%%%%%%%%%%%%%%%%%%%
% CONCLUSION
%
%
\section{Discussion}\label{sec:discussion}
The current study examined how social interaction in robotic agents dynamically changes depending on how the complexity in the free energy is regulated. For this purpose, we conducted simulation experiments on dyadic imitative interactions using humanoid robots equipped with PV-RNN architectures. PV-RNN is a hierarchically organized variational RNN model that employs a framework of predictive coding and active inference based on the free energy principle.
In a preparatory analysis we showed that PV-RNNs trained with looser regulation of complexity develop stronger action intentions by self-organizing more deterministic dynamics with strong initial sensitivity. Networks trained with tighter regulation develop weaker intentions by self-organizing more stochastic dynamics.\par
Our experiments revealed different types of interactions between robots. In the experiment where a robot having looser regulation interacts with a robot with tighter regulation, the former tends to lead the interaction by exerting action intention with stronger belief, while the latter tends to follow the other. The following robot adapts its posterior to its observations of the leading robot. In this setting, the synchronization of movement {\sffamily\textbf{B}} and {\sffamily\textbf{C}} ({\sffamily\textbf{B}}{\sffamily\textbf{C}}-synchronization rate) between the two robots was significantly higher than the chance rate. When two robots with looser regulation, i.e. intentions with stronger belief, interact, each robot tends to generate its own intended movements.
Finally, in case both robots have tighter regulation, a fluctuating dyadic interaction develops where each robot attempts to adapt to the counterpart with an intention with weaker belief. 
In the last two cases, the {\sffamily\textbf{B}}{\sffamily\textbf{C}}-synchronization rate was not significantly higher than the chance rate.
It can be summarized that the dyadic imitative interaction, including situations where the other's movements are unpredictable, tends to be synchronized successfully when a dedicated leader and follower are determined; a leader develops action intentions with strong belief whereas a follower develops action intentions with weak belief.\par 
The readers may ask why tighter or looser regulation of the complexity term results in development of weaker or stronger belief of action intention for each robot.
Let us consider a situation in which the PV-RNN learns to predict probabilistic sequences $\bm{\bar{X}}_{1:T}$ with meta-prior $\mathbf{w}$ set either with a large value (loose regulation) or a smaller one (tight regulation). 
The learning process infers the posterior mean $\bm{\mu}^q_t$ and standard deviation $\bm{\sigma}^q_t$ at each time step $t$. In order to minimize the error $\mathbf{e}$ in the accuracy term, $\bm{\mu}^q_t$ is fitted with an arbitrary value, where $\bm{\sigma}^q_t$ will be minimized, in both cases.
Notably, when the data ${\bm{\bar{X}}_t}$ is observed as random, the corresponding posterior $\bm{\mu}^q_t$ also becomes random.\par
Let us consider the two cases when the meta-prior $\mathbf{w}$ is either set large or small. In case $\mathbf{w}$ is set large, the KL Divergence between the posterior and the prior is strongly minimized. Thus, $\bm{\mu}^p_t$ and $\bm{\sigma}^p_t$ of the prior latent state become close to $\bm{\mu}^q_t$ and $\bm{\sigma}^q_t$ of the posterior. 
By this, $\bm{\sigma}^p_t$ in the prior is forced to take a minimal value close to $0$; therefore, the prior generation becomes deterministic.
Since $\bm{\mu}^p_{1:T}$ should be reconstructed as close to the sequence $\bm{\mu}^q_{1:T}$ inferred with randomness, the prior generative model is forced to develop strongly nonlinear deterministic dynamics with the initial sensitivity through learning.
On the other hand, if $\mathbf{w}$ is set with a small value, the KL Divergence is only weakly minimized. In this case, prior $\bm{\mu}^p_t$ and $\bm{\sigma}^p_t$ can diverge from the posterior ones; therefore, the learning becomes "relaxed".
As a result, the prior generative model develops stochastic dynamics with only weak non-linearity, wherein $\bm{\mu}^p_t$ takes an average of $\bm{\mu}^q_t$ over all occurrences and $\bm{\sigma}^p_t$ takes their distribution at each time step. Consequently, with larger $\mathbf{w}$, the generative model develops action intention with stronger belief (i.e. smaller $\bm{\sigma}^p$) whereas in the case of tighter regulation using a smaller $\mathbf{w}$, the generative model develops action intention with weaker belief (i.e. larger $\bm{\sigma}^p$).\par
The current experiments consider a fixed meta-prior setting only. Since the meta-prior is the essential network parameter to guide the strength of action intention in the proposed framework, future studies should target meta-learning of the meta-prior in developmental processes or through autonomous adaption within dyadic contexts. This could provide further understanding of more complex social interaction phenomena, including turn-taking in the context of adaptive regulation of the complexity term in free energy. 
%%%%%%%%%%%%%%%%%%%%%%%%%%%%%%%%%%%%%%%%%%%%%%%%%%%%%%%%%%%%%%%%%%%%%%%%%%%%%%%%
\section*{ACKNOWLEDGMENT}
We thank all members of the Cognitive Neurorobotics Research Unit. Special thanks goes to Fabien Benuerau and Jeffrey Queisser, for fruitful discussions about developing the computational model. The authors also acknowledge the support of the Scientific Computation and Data Analysis section of OIST to carry out the research presented here.
%%%%%%%%%%%%%%%%%%%%%%%%%%%%%%%%%%%%%%%%%%%%%%%%%%%%%%%%%%%%%%%%%%%%%%%%%%%%%%%%
%
% \bibliographystyle{IEEEtran}
% \bibliography{references.bib}

%
\end{document}